# TinyML: Analysis of Xtensa LX6 microprocessor for Neural Network Applications by ESP32 SoC

Md Ziaul Haque Zim
[1] *Department of Computer Science and Engineering*
*Daffodil International University*
Dhaka, Bangladesh
[2] *Department of Computer Science and Engineering*
*Saint Petersburg Electrotechnical University "LETI"*
Saint Petersburg, Russia
ziaul15-1133@diu.edu.bd

*Abstract*—In recent decades, Machine Learning (ML) has become extremely important for many computing applications. The pervasiveness of ultra-low-power embedded devices such as ESP32 or ESP32 Cam with tiny Machine Learning (tinyML) applications will enable the mass proliferation of Artificial Intelligent powered Embedded IoT Devices. In the last few years, the microcontroller device (Espressif ESP32) became powerful enough to be used for small/tiny machine learning (tinyML) tasks. The ease of use of platforms like Arduino IDE, MicroPython and TensorFlow Lite (TF) with tinyML application make it an indispensable topic of research for mobile robotics, modern computer science and electrical engineering. The goal of this paper is to analyze the speed of the Xtensa dual core 32-bit LX6 microprocessor by running a neural network application. The different number of inputs (9, 36, 144 and 576) inputted through the different number of neurons in neural networks with one and two hidden layers. Xtensa LX6 microprocessor has been analyzed because it comes inside with Espressif ESP32 and ESP32 Cam which are very easy to use, plug and play IoT device. In this paper speed of the Xtensa LX6 microprocessor in feed-forward mode has been analyzed.

*Keywords— TinyML, Xtensa LX6 microprocessor, Machine Learning, Neural Network, Embedded IoT Device, Espressif ESP32 and ESP32 Cam.*

## I. INTRODUCTION

The explosive growth in machine learning and miniaturization of electronics has been created new research opportunities. The thought of eco-friendly energy, which stresses the utilization of electric-driven gear [1,2], which has been recently in the discussion [3]. Machine Learning already achieved the great success with the multiple core enabled processor, and in recent days researchers thinking to apply small or tiny machine learning application on system on chip (SoC) microcontrollers with less processing power [4].

However, the extensive deployment of the machine learning algorithms will bring a clear rise in the artificial intelligence where a huge amount of processing power will be used due to process small algorithm with high-end multicore processor. Other animals and humans have a brain and a central nervous system to avail the process of information with neural networks. In the different fields, neural network has been successfully implemented, from detecting criminal [5,6] and medical diagnosis [7,8] through image classification [9], to autonomous driving [10].

As part of a long-established custom, practice, or belief, the training of neural network is commonly centralized, and for the computational intensity and big memory requirements it is mostly done on cloud servers. An instance of a particular situation the dataset of deep neural networks it could be more than several dozen terabytes for learning [11,12].

In centralized approach some researcher found disadvantages like data privacy issues [13], between devices and cloud it increased latency and network traffic for that forbid in some determination [14]. It also decreases the reliability, when exchanging of information is important between cloud and device or sensors [15]. And some researcher presents that sometimes the center can be SoC, smartphone and micro data center between source and data [16].

In the most recent decade, many researcher, engineers and professionals have studied on machine learning applications and implement these on microcontrollers, SoC's and embedded IoT devices. Special libraries [17] and hardware development with ML/AI functions and capabilities [18-20]. For the ESP32 and ESP32-S Series SoCs Espressif developed an IoT Development Framework named ESP-IDF [21] and ESP32 chip-based platform that is used for facial recognition and detection [22].

In addition, many research works have been done, studied and used ESP32 widely like in embedded IoT devices such as LPG gas leakage [23], home monitoring system [24], solar cell current and its battery power prediction [25] and air pollution detect system [26].

In this paper, the Tensilica Xtensa LX6 microprocessor by Cadence is discussed and its neural network capabilities will be analyzed [27].

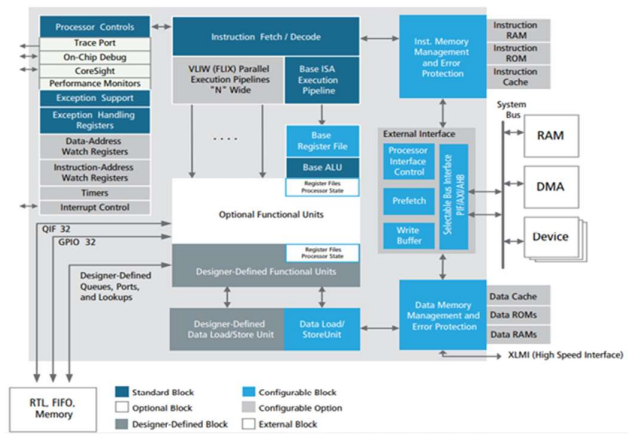

**Fig. 1.** Xtensa LX6 DPU showing standard, optional, and designer-defined blocks

In **Fig. 1**, the ESP32 SoC's processor Xtensa LX6 DPU showing standard, optional, and designer-defined blocks. For neural network acceleration, this SoC does not have any special hardware but some ML frameworks and examples for ESP32 chip has been developed by Espressif. So that I choose a ESP32 WROOM IoT development board by Espressif for the neural network acceleration that inside have a Xtensa LX6 microprocessor SoC.

A SoC or a new class of programmable processor that combines high-performance and industry-standard, software-programmable multi-core CPU is called data plane processing units (DPUs). Xtensa LX6 called as DPU. It's highly efficient, small and it have low-power 32-bit base architecture. In Fig. 2 an functional block diagram by Espressif is showed for ESP32.

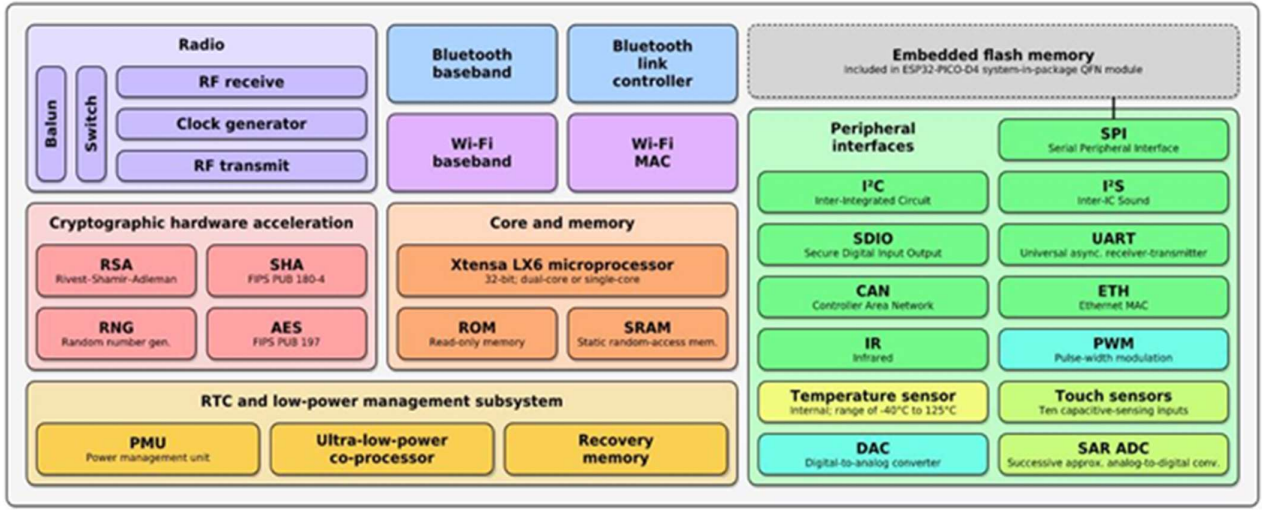

**Fig. 2.** Espressif Functional Block Diagram for ESP32 chip.

The formation of this study as follows. In section II, a scheme and machine learning usage on ESP32 is discussed previous studies. In section III, some study that deals with neural networks and Xtensa LX6 multi-core typical example or pattern will be described. In section IV, analysis of EPS32 processing unit Xtensa LX6's one, two and both cores speed of feedforward process propagation of neural network. The testing is done and finally in section V result will be analyzed and discussed.

## II. THE SCHEME

For analysis of Xtensa LX6 microprocessor for neural network applications, need to perform few necessary steps. An ESP32 development board is essential tools for this. Arduino IDE with Espressif ESP32 board compatible is important. Using Arduino IDE, a neural network will be developed and will implement into an ESP32 for the analyzing its processing power.

It can be observed that Xtensa LX6 microprocessor most are used for the general, including a few neurons. But there is no study report regarding machine learning applications that study, being researched or investigate the speed of running neural networks on a microcontroller in the real-world conditions.

In the next, the different approaches analyzed that already studied by the researchers.

### A. Machine Learning on the Cloud

By Chand et al. an ultrasonic sensor with ESP32 has been used to predict the behavior and analyzing the status of a person in a room. With the help of sensor, the ESP32 is used for collecting the data and send it to the cloud. A machine learning algorithm is used to analyze data and give decision whether the situation is normal or abnormal situation [28].

By Fernoaga et al an ESP32 camera has been used for take the images for the neural network's recognition [29].

Rosato and Masciadri has make a observation system by using an ESP32 that is non-invasive and detect the siting people by analyzing the data on cloud and the classification of the problem has been solved by Logistic regression algorithm [30].

By Islam et al. designed a module of biosensing with an Analogue Device AD8232, ESP32 as microcontroller to transmit data. A Linux server receive those data and analyze with neural network [31].

By Komarek et al has been presented that MQTT is used in abstraction layer and interface nodes with sensors. Those particular nodes are capable for transmit and receive data to another nodes. A cloud server is used for the analysis [32].

By Zidek et al. optimize data for IoT and accumulate them to the packet and transmit them to the cloud. These packets have multi-value [33].

### B. Machine Learning on Xtensa LX6 microprocessor by using ESP32

Very few authors has been studied for ESP32 for some Machine Learning applications.

Kokoulin et al implement a video recognition system that detects the presence of a face or silhouette fragment by using ESP32. For shorten the network traffic and computational load of facial recognition server and got the result of 80% to 90% traffic decrease [34].

Espressif Systems has been developed an open-source ESP-WHO framework that is available on GitHub for recognition and detection [22]. The Cascaded Convolutional

Networks model and new mobile architecture – Mobile Net V2 is based on this ESP-WHO framework [35,36].

*C. Machine Learning on Arduino with Arduino IDE*

Arduino is an open-source hardware platform and it have many hardware devices that being used for machine learning applications. Some author already implements machine learning applications on Arduino development boards by using Arduino IDE.

Neural network algorithm named C-Mantec that can add new neurons in the leaning process has been implemented by Ortega-Zamorano et al. [37]. Two hidden and one output neuron has been used to control PID controller and DC motor in Arduino Uno based development board, designed and implemented by Rai and Rai [38]. Adhitya et al. used neural networks with two inputs executed from sensor data readings. Whereas, two hidden layers with seven neurons each and two neurons as output layer [39].

### III. DECENTRALIZED NEURAL NETWORKS

In this study, the speed of single-machine parallelism in data propagation is tested in neural network feed-forward process. In the hidden layer by different number of neurons in Xtensa LX6 microprocessor has been done testing with it's one and two cores that was formerly mentioned.

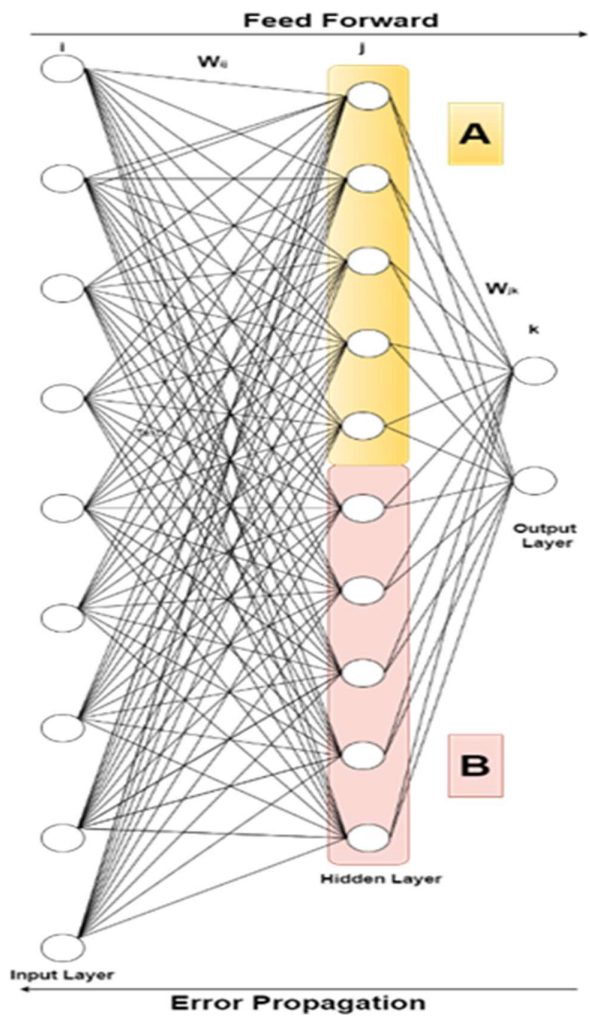

**Fig. 3.** Task Division of Hidden Layer

The process will be divided into two same-size groups for data propagation. Simple tasks division can be seen in the Fig. 3 by three method. For the upper half (A) the "core 0" will calculate values and for the lower half (B) another "core 1" will calculate values. After calculating the total (j = A+B) neurons of hidden layer a result will generate (j).

### IV. THE FEED FORWARD PROCESS AND THE PROPAGATION SPEED

*A. Hardware Preparation*

For testing Xtensa LX6 microprocessor propagation speed of neural network process with it both cores, an ESP32 SoC is used. In the market, there are various type of ESP32 boards are available and for the testing purpose "ESP-WROOM-32" ESP32 board has been used that is produced by Espressif and showed in Fig. 4.

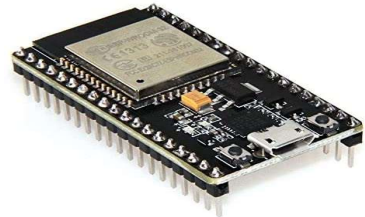

**Fig. 4.** "ESP-WROOM-32" ESP32 Development Board

*B. Software Preparation*

Because of opensource there are various way to program ESP32 boards. Two main method for programming ESP32:

  a. Espressif IoT Development Framework "ESP-IDF"
  b. Arduino IDE, ESP32 Arduino Core

Compared with ESP-IDF and Arduino IDE. ESP-IDF low-level programming whereas Arduino IDE has some extra facilities. In this study, for testing Xtensa LX6 microprocessor Arduino IDE is used to measure propagation speed of this process for this feedforward neural network.

*C. Measuring the Execution time*

For measuring the execution time there are various way. In this study, micros() the simple function is used.

**Example Code:**

```
time_start = micros();
//some code
time_end = micros();
time = time_end – time_start;
```

Arduino IDE receive the values through serial monitor when micros() function returns the number of microseconds [40].

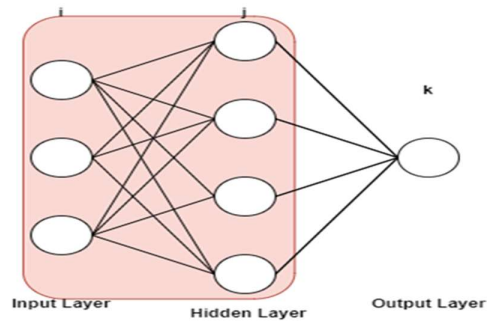

**Fig. 5.** Pre-activation Layer

In Fig. 5, it can be seen which part is being implemented. In fact, this part is time dependent for measuring in neural networks with each hidden layer. The number of inputs is much higher than the normal neural outputs. Expressed as following equation,

$$z = b + \sum_{i=1}^{N} a_i w_i \qquad (1)$$

Whereas, $b = Bias$ , $w_i = weights$ , $a_i = $ input values , $z = a$ sum of all multiplications of $a_i$

*D. Experiments and Results*

Xtensa LX6 microprocessor on ESP32 board have two cores and for the first time only one core is used for measuring the propagation time. Describes in TABLE I. The FreeRTOS have xTaskCreatePinnedToCore() function which is capable for setting the specific processor core for task. In ESP32 have only two cores and for the first segment of taking measurements with only one core when the function is set to 0. Last argument in the function is set to 1, this argument can be "core 0" or "core 1".

TABLE I. RESULTS WITH SINGLE CORE IN XTENSA LX6 MICROPROCESSOR ON ESP32

| No | Layer | | Operations | Time |
|---|---|---|---|---|
| | Input | Hidden | | |
| 1 | 50 | 20 | 1000 | 383 $\mu s$ |
| 2 | | 40 | 2000 | 681 $\mu s$ |
| 3 | | 60 | 3000 | 966 $\mu s$ |
| 4 | | 80 | 4000 | 1258 $\mu s$ |
| 5 | | 100 | 5000 | 1582 $\mu s$ |
| 6 | | 120 | 6000 | 1830 $\mu s$ |
| 7 | | 140 | 7000 | 2203 $\mu s$ |
| 8 | | 160 | 8000 | 2415 $\mu s$ |
| 9 | | 180 | 9000 | 2815 $\mu s$ |
| 10 | | 200 | 10000 | 3071 $\mu s$ |

Some additional setup and code are needed for measuring the propagation time in Xtensa LX6 microprocessor on ESP32. For single core, all task has been done in using only one core. But for using both cores, as I mentioned previously Xtensa LX6 microprocessor have two cores, all tasks have to be divided into two parts.

TABLE II. describe about the results of using two cores in Xtensa LX6 microprocessor on ESP32. In this study, for simplicity, all the neurons in the hidden layer setup multiplied by 20 times, even in numbers and loops. Half of the neuron is counted at processor core 0 and the other half of the neuron is counted at processor core 1 and Fig. 3. gives a crystal-clear idea about this process. Whereas, core 0 is the first core and core 1 is the second core of Xtensa LX6 microprocessor.

The aforementioned function xTaskCreatePinnedToCore is called twice to select both processor core setup. When the first time it is called, it gets the value of the first half of the neuron in the hidden layers.

Assume that, there are 20 neurons in hidden layers and the written function input parameter will get values 0 to 9 in the first call and 10 to 19 values will take in second call as input parameter. A global array is declared for define inputs and weights values, because every core need access on them.

**Example Code:**

```
xTaskCreatePinnedToCore(
    powerTask,      /* Function to implement the task */
    "powerTask",    /* Name of the task */
    1000,           /* Stack size in words */
    (void*)&twoTasks1,  /* Task input parameter */
    20,             /* Priority of the task */
    NULL,           /* Task handle. */
    0);             /* Core where the task should run */
xTaskCreatePinnedToCore(
    powerTask,      /* Function to implement the task */
    "coreTask",     /* Name of the task */
    1000,           /* Stack size in words */
    (void*)&twoTasks2,  /* Task input parameter */
    20,             /* Priority of the task */
    NULL,           /* Task handle. */
    1);             /* Core where the task should run */
```

TABLE II. RESULTS WITH BOTH CORE IN XTENSA LX6 MICROPROCESSOR ON ESP32

| No | Layer | | Operations | Time |
|---|---|---|---|---|
| | Input | Hidden | | |
| 1 | 50 | 20 | 1000 | 253 $\mu s$ |
| 2 | | 40 | 2000 | 406 $\mu s$ |
| 3 | | 60 | 3000 | 547 $\mu s$ |
| 4 | | 80 | 4000 | 699 $\mu s$ |
| 5 | | 100 | 5000 | 858 $\mu s$ |
| 6 | | 120 | 6000 | 976 $\mu s$ |
| 7 | | 140 | 7000 | 1159 $\mu s$ |
| 8 | | 160 | 8000 | 1268 $\mu s$ |
| 9 | | 180 | 9000 | 1482 $\mu s$ |
| 10 | | 200 | 10000 | 1599 $\mu s$ |

Elapsed time ration can be seen in TABLE III. For the number of multiplication and addition operations on Xtensa LX6 microprocessor cores on ESP32 boards, it is obvious that ration is rising. If increasing number of operations has less than the impact is less.

TABLE III. COMPARISON RESULTS OF XTENSA LX6 MICROPROCESSOR CORES ON ESP32

| No | Time (in $\mu s$) | | Ratio |
|---|---|---|---|
| | One Core | Two Core | |
| 1 | 383 | 253 | 1.51 |
| 2 | 681 | 406 | 1.68 |
| 3 | 966 | 547 | 1.76 |
| 4 | 1258 | 699 | 1.80 |
| 5 | 1582 | 858 | 1.84 |
| 6 | 1830 | 976 | 1.87 |
| 7 | 2203 | 1159 | 1.90 |
| 8 | 2415 | 1268 | 1.90 |
| 9 | 2815 | 1482 | 1.90 |
| 10 | 3071 | 1599 | 1.92 |

In these experiments, the time unit is calculated in microsecond *($\mu s$)*. It's equal to one millionth (0.000001 or $10^{-6}$ or $\frac{1}{1,000,000}$) of a second.

## V. DISCUSSION AND CONCLUSION

Xtensa LX6 microprocessor have two cores on ESP32. The main theme of Amdahl's law is the original execution time divided by an enhanced execution time [41]. In the recent decades, the modern version of Amdahl's law says that enhancing the fraction $f$ of a computation by a speedup $S$, then I get [41],

$$Speedup_{enhanced}(f,S) = \frac{1}{(1-f)+\frac{1}{S}} \quad (2)$$

In the Fig. 6, we can see a chart of comparison results of Xtensa LX6 microprocessor cores on ESP32 from the TABLE III.

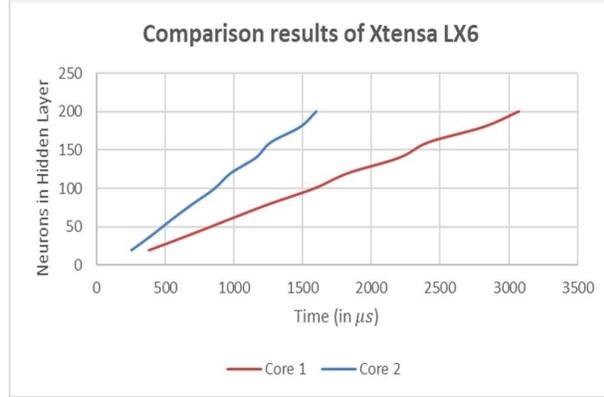

**Fig. 6.** Comparison results of Xtensa LX6 cores on ESP32

In this study, Xtensa LX6 microprocessor have two cores and from Amdahl's law perspective the result will be analyzed. So that I formulated the Amdahl's law as,

$$S_{latency}(S) = \frac{1}{(1-p)+\frac{p}{S}} \quad (3)$$

Whereas, $S_{latency}$ is code execution speedup, $S$ is instead of value 2 because of Xtensa LX6 microprocessor multiple cores, and $p$ is code execution time proportion.

According to this equation (3) in TABLE IV, the values for $p$ are calculated.

TABLE IV.  PROPORTION OF THE CODE EXECUTION TIME

| Ratio | Number of Operations | $p$ |
|---|---|---|
| 1.51 | 1000 | 0.68 |
| 1.68 | 2000 | 0.81 |
| 1.76 | 3000 | 0.86 |
| 1.80 | 4000 | 0.89 |
| 1.84 | 5000 | 0.91 |
| 1.87 | 6000 | 0.93 |
| 1.90 | 7000 | 0.95 |
| 1.90 | 8000 | 0.95 |
| 1.90 | 9000 | 0.95 |
| 1.92 | 10000 | 0.96 |

It is noticeable in the TABLE IV, that the calculated values of $p$ proportion of code execution time is increased and this is obvious.

This is because data transfer between different types of memory has less and less impact on the total run time. If and only if static tasks for processor cores setup [4].